\documentclass[11pt]{article}
\usepackage{paperstyle}
\usepackage{url}
\usepackage{xcolor}
\usepackage{graphicx}
\usepackage{tabularx}
\usepackage{subcaption}

\usepackage{array}
\usepackage{multirow}
\usepackage{tabularx}
\usepackage{booktabs}
\begin{document}
\title{An Approach to Intelligent Pneumonia Detection and Integration}
%
%

\author{
  Bonaventure F. P. Dossou \\
 Jacobs University Bremen\\
\texttt{fdossou@jacobs-university.de}
  \\\And
  Alena Iureva\\
  Jacobs University Bremen\\
  \texttt{aiureva@jacobs-university.de}
  \\\AND
  Sayali R. Rajhans\\
  Jacobs University Bremen\\
  \texttt{srajhans@jacobs-university.de}
  \\\And
  Vamsi Sai Pidikiti\\
  Jacobs University Bremen\\
  \texttt{vpidikiti@jacobs-university.de}
}

\maketitle

\begin{abstract}
Each year, over 2.5 million people, most of them in developed countries, die from pneumonia \cite{1}. Since many studies have proved pneumonia is successfully treatable when timely and correctly diagnosed, many of diagnosis aids have been developed, with AI-based methods achieving high accuracies \cite{2}. However, currently, the usage of AI in pneumonia detection is limited, in particular, due to challenges in generalizing a locally achieved result. In this report, we propose a roadmap for creating and integrating a system that attempts to solve this challenge. We also address various technical, legal, ethical, and logistical issues, with a blueprint of possible solutions.
\end{abstract}

\section{Motivation}
According to the World Health Organization (WHO), pneumonia is a treatable disease that causes millions of people to die from lack of access to its prevention and treatment, including over 800 thousand children under five every year. The problem is perpetuated by mistakes in diagnosis, causing interventions to be too late and leading to death. Caused by bacteria or viruses, pneumonia is worse for populations harmed by pollution and can be prevented by vaccination or treated with antibiotics \cite{3}. One of the most popular ways to detect pneumonia is Chest X-ray (CXR); others include blood tests, pulse oximetry, and Chest CT scans \cite{4}.

The technological progress and the need to address the shortage of medical opportunities in some geographical regions have recently given growth to various telehealth solutions, including teleradiology. Besides, AI is increasingly being used hand-in-hand with telehealth technologies to reduce the workload on healthcare workers. The tasks that it attempts to perform include patient monitoring, intelligent assistance, diagnostics, remote collaboration, and self-diagnosis tools, including health trackers \cite{5}. By 2017, 86\% of major healthcare companies used some AI in their operations, thus reducing operational costs by 10\% \cite{6}.

In general, studies suggest that AI-based diagnosis systems, in particular, could assist clinicians by achieving fast, reliable data interpretation, improve and streamline workflow for health systems, achieving faster and seamless processing \cite{7}, reduce medical errors (\cite{7}, \cite{8}), help confirm a diagnosis or clarify it \cite{9}. Besides, certain implementations can empower patients by providing them more in-depth but understandable information about their health.

While there have been some progress in automated pneumonia diagnosis by CXR using Deep Convolutional Neural Networks (CNNs), as we will describe in section \ref{model}, there are multiple challenges to tackle before the resulting models become genuinely production-ready. Some of them are related to computational power, but the most considerable technical challenge is insufficient data — often caused by the fact that experimental systems are often developed and tested within single hospital’s walls. To build a diagnosis system applicable at scale, one needs a dataset diverse by geography, demographic characteristics, scanner brand, and other variables \cite{10}. Therefore, we find it appropriate to try and address this issue in this work.

Ultimately, if successfully implemented, an automated pneumonia diagnosis system could save many lives by saving radiologists’ time. CXR is a relatively cheap and fast diagnostic method, with rising demand, yet two-thirds of the world population do not have access to radiologists \cite{11}. Nevertheless, even the remaining third gets 1.44 billion CXRs annually \cite{12}. Therefore, it is clear that even a 10\% improvement in CXR reporting speed, let alone expansion of the access, could add up to tens and hundreds of thousands of saved lives annually.

\section{Background and Related Work}
As there is by now much research dedicated to AI-based pneumonia detection by CXR, let’s take a look at a few examples.
\begin{itemize}
    \item Rajpurkar et. al., in their paper "CheXNet: Radiologist-Level Pneumonia Detection on Chest X-Rays with Deep Learning", using the ChestX-ray14 dataset released by (Wang et al., 2017) which contains 112,120 frontal-view X-ray images of 30,805 unique patients, achieved AUC of 76,80\% with a F1-Score of 43.50\%. Their model performed better than the models proposed respectively by (Wang et al., 2017) and (Yao et al., 2017) \cite{13}.
    \item Researchers from India compared different combinations of CNNs as feature extractors and simple ML models as classifiers on a set of 112k CXR images resized to 224x224 and found the combination of DenseNet-169 + Support Vector Machine (SVM) to be the best within the scope of their study, achieving AUC of 80\% \cite{14}.
\end{itemize}

Overall, a systematic literature review on intelligent pneumonia detection methods \cite{15}, points that CXRs are often subject to complex preprocessing, including morphological operation, edge detection, and Gaussian filters. The resulting features are used in classification via Neural Networks (NNs), SVM, Random Forest Classifier (RFC) or another model or classifier. As for NNs architectures, many, including VGGNet (\cite{13}, \cite{14}, \cite{16}), AlexNet, LeNet, GoogleNet \cite{16}, ResNet (\cite{14}, \cite{16}), DenseNet, and Xception \cite{14}, were tried, with decent results. But it worths emphasizing that models with too many layers may be impractical due to excessive computation time. Performance metrics include accuracy, precision, recall, F1 score, and AUC \cite{15}, yet for medical applications, recall is often the most critical metric due to the dire consequences of false negatives in diagnostics \cite{2}.

\section{Methodology}
\subsection{Application architecture}
For the sake of applicability in under-served areas, the solution has to fulfill the following requirements:
\begin{itemize}
\item Can function under low network speeds efficiently, even for dial-up (up to 56 kbit/s), which can still be the case in rural areas of some countries \cite{17}.
\item Can function autonomously if the internet connection is off
\item Does not require extensive network usage on the client-side to reduce costs
\item Is fast enough to reduce the time spent to evaluate a CXR (which usually takes up to 15 minutes) \cite{18}.
\item Can provide the system (server-side) with an ongoing supply of additional data to increase performance
\end{itemize}

\begin{figure}[t]
\includegraphics[width=\linewidth]{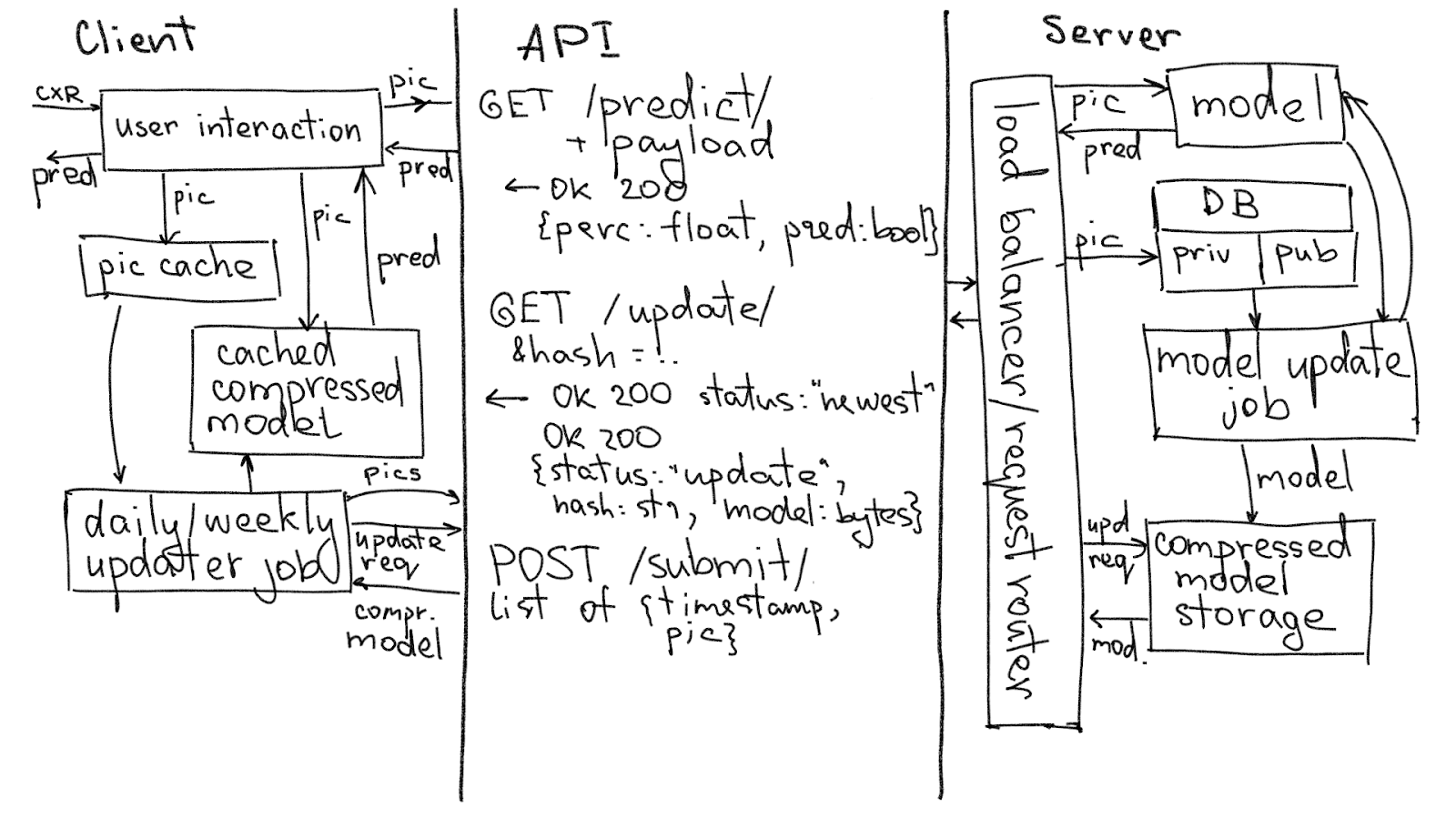}
\centering
\caption{\label{fig0} Application Architecture and Workflow Diagram}
\end{figure}

The Figure \ref{fig0} presents the workflow diagram of the application. Here are some key factors it worths knowing:
\begin{itemize}
\item The client interacts with the user (presumably a radiologist/technician) in a simple (and single) way: the user uploads a CXR and receives the probability of the picture indicating pneumonia and the model’s binary verdict. The user will also be able to mark whether the diagnosis was confirmed or not to gather the data.

\item The CXR is preprocessed on the client, thus reducing the picture size to 4-16KB, uploadable on a server within a few seconds even with dial-up, assuming 8-bit color depth grayscale 128x128 image.

\item Usually, a prediction request is sent to the server. Because the response is two numbers with labels, its size is negligible in comparison to headers. If the request was unsuccessful or the internet connection is not available, prediction happens locally, and the preprocessed picture is put to the picture cache to send later.
\item While neural networks can be large, for example being 240MB for AlexNet, the combination of Pruning, Quantization, and Huffman encoding can reduce these numbers by 30-50x depending on the specific architecture and weights while speeding up 3-4x layerwise and retaining accuracy \cite{19}. This finding means that AlexNet would reduce to 6.9MB, downloadable in 16 minutes at 56 Kbit/sec (long, but bearable) — and because we initially propose a simpler model, the actual time would be even less.
\item The full (uncompressed) network would run on the server, and the compressed version of the most recent network could be downloaded for autonomous processing. The update will only happen if the hash sum of the client’s current model is different from the hash sum of the newest model.
\item The overall weekly network usage, assuming 100 scans a day, a weekly update, 1-kilobyte headers, and 5MB compressed NN size, would be up to 100*(17+1)*7+5*1024=17720KB, or under 18MB, making it doable even for remote rural areas in developing countries.
\item The processed pictures in the database would be split into already used in the model, and the update batch material, and a public and private section. The section can be chosen depending on the specific contractor preferences regarding sharing the anonymized personal data. The public section is to form publicly available datasets for comparison of AI solutions, while private only for model training.
\item Once in a while, the model would be subject to a transfer learning process with new data and replace the older model if the newer one performs better.
\end{itemize}

\subsection{Data Processing, Model's Architecture and Training}
\label{model}
For the model's creation, we illustrate the process with a particular Kaggle dataset of 5856 images \cite{20}.
\subsubsection{Data Processing}
Each picture of the dataset is a grayscale picture with resolution 1152x760. To every image, we apply the gamma correction, resulting in correcting the brightness of input image. This is done by using a non linear transformation between the input values and the mapped output values:
\begin{center}
$O = (\frac{I}{255})^\gamma \times 255$    
\end{center}
\begin{figure}[!htb]
\centering
\begin{minipage}{.5\textwidth}
  \centering
  \includegraphics[width=\linewidth]{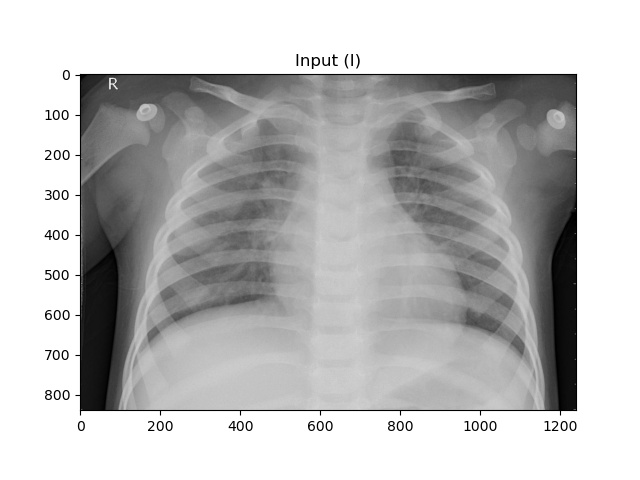}
\end{minipage}%
\begin{minipage}{.5\textwidth}
  \centering
  \includegraphics[width=\linewidth]{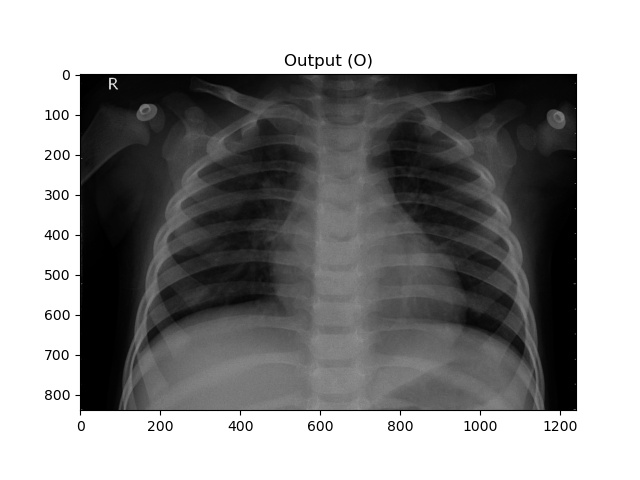}
\end{minipage}
\caption{\label{fig1} Transformation of a chest X-ray from a patient suffering from Pneumonia, with $\gamma=2.8$}
\end{figure}

\begin{figure}[!htb]
\centering
\begin{minipage}{.5\textwidth}
  \centering
  \includegraphics[width=\linewidth]{input.jpg}
\end{minipage}%
\begin{minipage}{.5\textwidth}
  \centering
  \includegraphics[width=\linewidth]{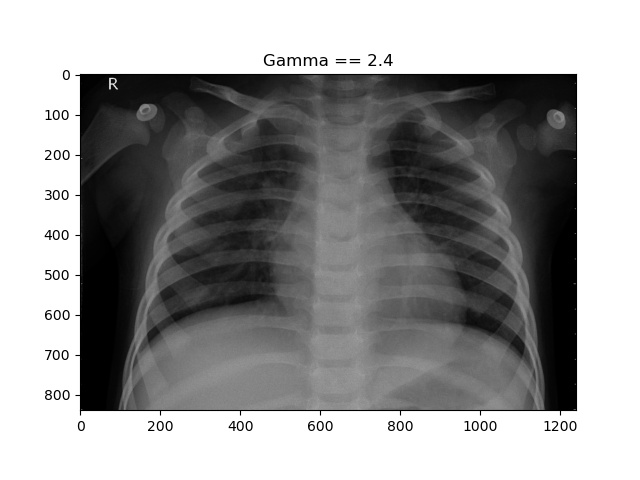}
\end{minipage}
\caption{\label{fig2} Transformation of a chest X-ray from a patient suffering from Pneumonia, with $\gamma=2.4$}
\end{figure}

\begin{figure}[!htb]
\centering
\begin{minipage}{.5\textwidth}
  \centering
  \includegraphics[width=\linewidth]{input.jpg}
\end{minipage}%
\begin{minipage}{.5\textwidth}
  \centering
  \includegraphics[width=\linewidth]{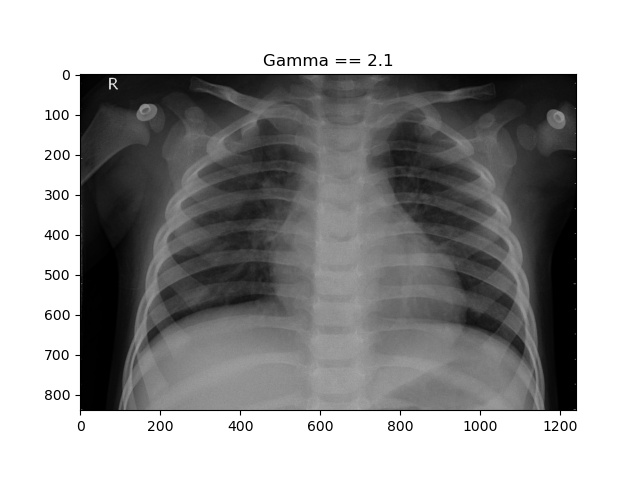}
\end{minipage}
\caption{\label{fig3} Transformation of a chest X-ray from a patient suffering from Pneumonia, with $\gamma=2.1$}
\end{figure}
where $O$ is the output image and $I$, the input image. Performing the Gamma correction, help our model to perform better. Figures \ref{fig1}, \ref{fig2} and \ref{fig3} show an example of the gamma correction, with $\gamma = 2.8, \gamma = 2.4$, and $\gamma = 2.1$. In practise, the larger $\gamma$, the dimmer is the picture. Lower values of $\gamma$ make images brighter and doesn't improve the performance of our model.
Further, for ease of processing, every image has been to 128x128. Along the tensor of training features, the scale the raw pixel intensities has been mapped to the range [0, 1]. As far as the training labels are concerned, $Normal$ has been mapped to 0 and $Pneumonia$ to 1. 
\subsubsection{Undersampling and Oversampling}
During the testing phase of our model, we have notice that the model almost perfectly classifies $Pneumonia$ but fails at classifying $Normal$ and predicts it in majority as $Pneumonia$. The dataset contains 5232 unique pictures, with only 25,75\% (1349) pictures of $Normal$. The gap is huge, and hence the model tends to more likely predict everything as $Pneumonia$. To overcome this barrier, we have oversampled the minority class $Normal$ and undersampled the majority class $Pneumonia$. Oversampling is defined as adding more copies of the minority class, along the dataset. Undersampled is defined as the opposite of oversampling. To perform $oversampling$ and $undersampling$, We have splitted our general data into test and train sets. Otherwise, the exact same observations could be present both in the test and train sets. This will make our model to memorize specific samples, and then cause overfitting and poor generalization to the test data.
\subsubsection{Data Augmentation}
Because there are so few samples initially, it is reasonable to perform data augmentation. This has been done using the ImageDataGenerator object of the deep learning framework Keras, with the parameters stated in Table \ref{imagegenerator}, resulting in 40992 training samples.
\begin{table}[]
\begin{center}
\begin{tabular}{ | m{7em} | m{1.5cm}| m{1.5cm} | } 
\hline
Parameters & Values\\ 
\hline
rotation\_range & 30\\ 
\hline
width\_shift range & 0.2\\ 
\hline
height\_shift range & 0.2\\
\hline
shear\_range & 0.2\\
\hline
zoom\_range & 0.2\\
\hline
horizontal\_flip & True\\
\hline
vertical\_flip & True\\
\hline
fill\_mode & nearest\\
\hline
\end{tabular}
\caption{\label{imagegenerator} Parameters of the ImageDataGenerator object for Data Augmentation}
\end{center}
\end{table}
\subsubsection{Model's Architecture and Training}
\begin{figure}[t]
\includegraphics[width=\linewidth]{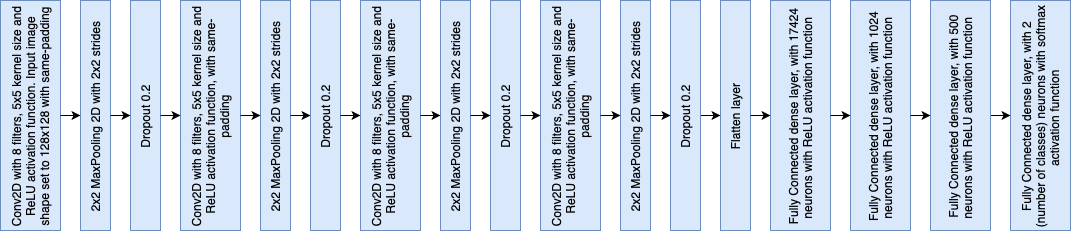}
\centering
\caption{\label{fig4} Model's Architecture}
\end{figure}

We opted for a very simple CNNs model, with relatively small number of layers. As future work and improvement, exploring Unet \cite{21} could be an option. The architecture of our model is described in Picture \ref{fig4}. For the training, we set the batch size to 64, the learning rate to 0.001 and the training epochs, initially at 500. For the loss function, we choose categorical entropy as the loss function, and as optimizer, we opted tried Stochastic Gradient Descent (SGD) and Adam, with a decay parameter set to 0.9. A callback has been also implemented, to evaluate the model after each epoch so that the weights are only saved when the current epoch’s validation loss, is less than the validation loss of the previous one, with patience parameter set to 100. Finally, to judge the performance, we use Accuracy, F1 and Recall metrics.
\begin{figure}[t]
\includegraphics[width=\linewidth]{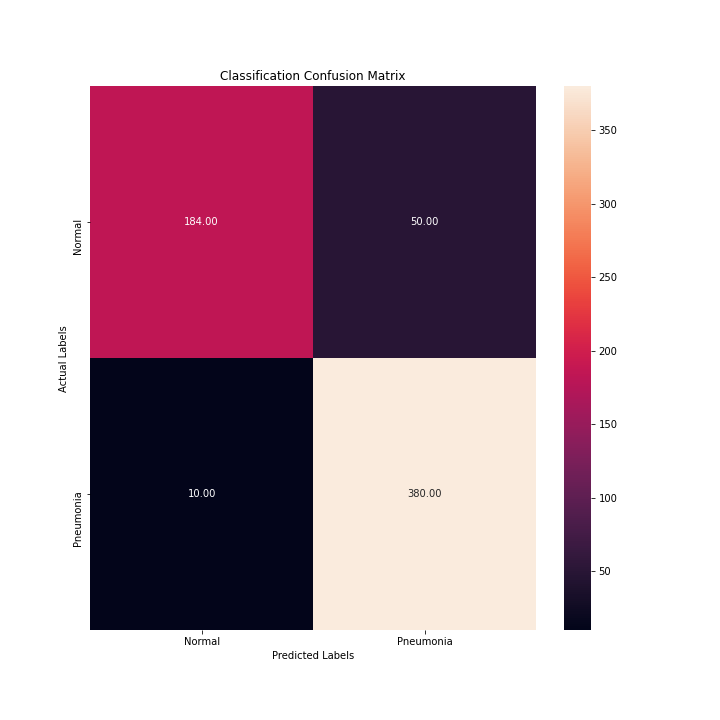}
\centering
\caption{\label{fig5} Model's Confusion Matrix}
\end{figure}

\section{Initial Results and Findings}
Our best result is achieved with Adam as optimizer, with an accuraccy of prediction of 91,04\%, and AUC of 88,04\%, on our test dataset, made of 624 unique samples. The Table \ref{results} presents the classification report of our model. The Figure \ref{fig5} presents the confusion of the classification task, performed by our model. We can from the confusion matrix, notice that the Pneumonia has been almost perfectly detected. Its is however not completely the case for $Normal$, for which the model has predicted 79\% of the samples correctly. This means that there is still headroom for improvements and perfection.

We want to end this section of our findings and results by sharing pairs of inputs and outputs (Figures \ref{fig5} and \ref{fig6}), with the heatmaps, indicating the parts of the inputs that helped our model to classify correctly.

\begin{figure}[!htb]
\centering
\begin{minipage}{.5\textwidth}
  \centering
  \includegraphics[width=\linewidth]{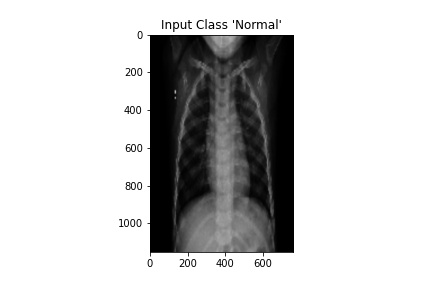}
\end{minipage}%
\begin{minipage}{.5\textwidth}
  \centering
  \includegraphics[width=\linewidth]{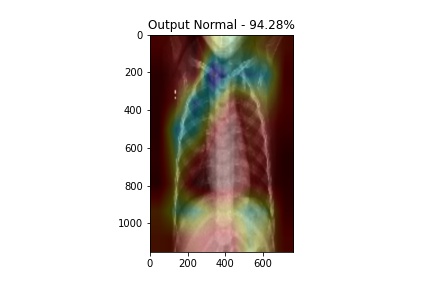}
\end{minipage}
\caption{\label{fig5} Heatmap Activations for $Normal$ class Detection}
\end{figure}

\begin{figure}[!htb]
\centering
\begin{minipage}{.5\textwidth}
  \centering
  \includegraphics[width=\linewidth]{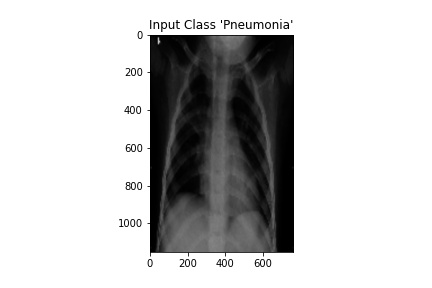}
\end{minipage}%
\begin{minipage}{.5\textwidth}
  \centering
  \includegraphics[width=\linewidth]{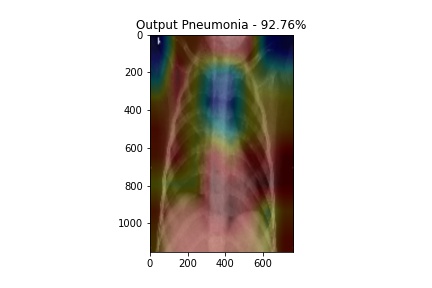}
\end{minipage}
\caption{\label{fig6} Heatmap Activations for $Pneumonia$ class Detection}
\end{figure}

\begin{table}[]
\resizebox{\columnwidth}{!}{
    \centering
    \begin{tabular}{lllll}
    \toprule
         & \textbf{Precision} & \textbf{Recall} & \textbf{F1-Score} & \textbf{Support}\\
    \midrule
        Normal & \textbf{95} & 79 & 86 & 234\\
        Pneumonia & 88 & \textbf{97} & \textbf{93} & 390\\
        \hline
        accuracy &  &  & \textbf{90} & 624\\
        macro avg & \textbf{92} & 88 & 89 & 624\\
        weighted avg & 91 & \textbf{90} & \textbf{90} & 624\\
    \bottomrule
    \end{tabular}
    }
    \caption{Classification Report of our Model}
    \label{results}    
\end{table}

\section{Technical challenges}
\subsubsection{Diversity of data}
As mentioned previously, to achieve broad applicability, the data should be diverse and representative in many respects. Under-representation of samples may cause the system to be less accurate for some patients, leading to increase in misdiagnosis. Diversity includes the hospital source, geographic location, demographic characteristics of patients like age, sex, ethnicity, income level, the X-ray scanner brand, and many others \cite{10}. To remedy this issue, one has to solve the problem of acquisition — complicated by data collection issues below. The representation problem is partially solved by selective oversampling and data augmentation, particularly by representative data collection strategies.

To help track and improve data diversity, the solution’s database can be augmented by diversity variables listed above for every CXR. Another improvement may be made by extracting a diverse subset from a large dataset for another iteration of model training \cite{22}.

Nevertheless, this challenge is never fully resolved, and significant improvements are only possible with increasing, possibly international reach and big data collection and cannot be fully tackled from day one based on a small set of open-source datasets.

\subsubsection{Intrinsic ML biases}
A lot of subtle mistakes are made by ML engineers every day throughout their work, which can be unnoticeable initially but make a considerable difference large-scale. These include dataset shift, unintended discriminatory bias, or accidental fitting of confounders \cite{23}. Biases entering models by mistake or due to sloppy design affect the right to equal treatment regardless of protected characteristics and tend to affect economically vulnerable populations significantly \cite{24}.

Similarly, as scientific papers are subject to peer review and code of software engineers to code review, open-sourcing most parts of the system can provide some oversight and reduce the risks. Commentary from independent experts should be welcome, and an opportunity for comparison to other AI models can increase the competition even further. An input of vulnerable and affected populations can also help discover problems previously unseen.

\subsubsection{Performance metrics clarity}
In our setup, the more meaningful metrics in the clinical setting are different from metrics that we use to optimize the model. An essential task in medicine is reducing false negatives or improving recall, while false positives are not that crucial. However, the easiest way to get a recall of 1 is to always detect pneumonia, yet such a model is useless. Because of that, more holistic measures are typically used, such as the F1 score.

The one thing that can be done quickly is outputting recall and precision for a current version of the model to the user. Another thing that acknowledges that precision should not be zero could be a weighted modification of the F1 score, where recall is given more weight. Precision should be decent, but recall must be high.

\subsubsection{Privacy and security issues}
Any technical solution should include robust security measures, yet it is even more accurate for those dealing with sensitive medical data. One should be careful with AI so that the model does not get tampered with via replacement in MiTM attacks or a newer concept of data diet attacks. Failure of defense from cyber attacks severely violates patients’ right to privacy, and the risks increase along with AI usage \cite{24}.

There is no easy way to tackle any of it. However, the bare minimum should be complex measures, including risk assessment, security protocols followed by involved parties, extensive testing and static analysis during software development, regular pen-testing, encryption, and security-driven network management. Besides, data collection from unconfirmed sources should not be permitted.

\subsubsection{AI systems comparison transparency}
As stated before, it is vital to compare different intelligent systems, yet it is currently complicated due to the unavailability of data and often different input formats. With time, given that there are incentives to open-source datasets and perform independent comparison studies, transparency may increase, but it is hardly possible to influence the process throughout just one project significantly.

\subsubsection{Modern AI’s subject matter unawareness}

The use of AI in diagnosis is currently limited due to the models’ general unawareness of human anatomy and physiology. This fact means it lacks expert knowledge to make a final decision, making it vital to have a human expert in using it as a tool \cite{10}. Therefore, it cannot reason on whether the diagnosis makes sense. While the first steps in knowledge networks are being done, increasing the system’s independence is beyond this report’s scope.

\section{Legal and ethical challenges}
\subsection{Data access and privacy regulations}
To remedy some of the existing privacy and human rights, extensive personal data regulations are being developed, EU’s GDPR \cite{25} being one of the most known legislations. While these laws define a baseline for privacy preservation, most often, the delineated guidelines are too complex to implement all at once, especially for a data-intensive product.

For the international targeting, especially considering that we are proposing the solution most beneficial for developing countries, many of them emerging democracies with weaker data protections, a reasonable way may be to create a roadmap of adoption based on a gradual increase of protections \cite{26} and working on compliance to more robust guidelines for targeting the new geographic area. At the same time, as possible, storing data primarily in locations providing more data protections is preferable to avoid problematic trans-border data transfers. While this approach may seem ethically dubious, in some cases, the need to ensure numerous people’s survival and health may temporarily outweigh privacy concerns.
\subsection{Lobbying for implementation or adoption}
To make an impact, widespread collaboration with healthcare officials and non-profit organizations is necessary. These players make decisions based on many potential impact points, weighing cost and benefit across many axes. Therefore, to accelerate the system’s widespread adoption, a convincing case should be made that the system is significantly beneficial.

It can be done based on the fact that at a small scale, it is possible to crowdfund or score a grant and implement the solution on an experimental basis. The further analysis of such experiments will build a ground for broader adoption in case of success, provided that a clear plan exists. The analysis may include monetary evaluation, labor analysis, patient outcomes monitoring, and an overview of intended and unintended consequences. 

\subsection{Implications of AI-caused medical errors}
Bias may be discriminative, but medical errors leading to deaths have another magnitude of responsibility \cite{23}. Depending on how liability is determined, the whole developer team may be held liable for a fatal mistake. Besides, accountability is hard to achieve for AI, as the specific decision algorithms are often unobvious — yet any error essentially violates the right to effective treatment and, in some cases, right to life \cite{24}.

At the project level, higher transparency and participation of medical specialists can reduce the risks. Risk prevention principles should also be developed and followed, and all parties informed of intrinsic AI shortcomings. These measures are far from enough, but more efficient measures should be implemented on the global, systemic level.
\section{Implementation challenges}
Following the intuition that the more usable the product is, the more it is used, it is known that usability explains up to 70\% of the variance in purchasing decisions \cite{27}. To facilitate the solution’s broader outreach, it is essential to wrap it into acceptable UX practices. The interface should be intuitive and straightforward, and small-cohort interviews with radiologists and technicians should be conducted in every location to improve the experience. In our case, good UX is incredibly vital, as it may help reduce human errors and directly lead to saved lives.
\subsection{The issue of trust}
Lack of trust both among medical professionals and patients may hinder the adoption of AI-based solutions directly and via the influence of government decisions \cite{23}. Despite the widespread concerns over lost jobs due to automation, majorities of people in selected developed countries, like India, Mexico, and Brazil, keep a high level of trust in both technology in general \cite{28} and AI in particular \cite{29}. Still, as more affluent countries tend to be less trusting, we can expect more distrust as countries develop economically.

Moreover, it’s one thing to trust abstract technology and another to trust its specific application. Concerns may be raised by old-school doctors and managers and those concerned whether such usage may violate any religious and spiritual principles. Therefore, it is essential to increase transparency, be honest about impacts, and educate the system’s direct users to increase trust. The trust issues in decision-makers can also be overcome via demonstration of significant material benefits.
\subsection{Material resources and skills availability}
Finally, some locations which would benefit most from our solution currently lack resources to access it. This may mean lack of internet connection \cite{30}, computers \cite{31}, or even electricity \cite{32}. Besides, the staff may not have any computer skills, which may also hinder progress.

While some usage training may be appropriate, it should not be challenging, so the user experience should be optimized for maximum simplicity. For this reason, and also should the unconnected field use be the goal, a dedicated battery-powered device as a client may be appropriate rather than a computer. It could also have a possibility of radio transmission to the reception points to compensate for the lack of internet or use satellite infrastructure, similarly to GPS devices.

\section{Acknowledgements}
This paper derives from our Big Data Semester project, led and supervised by the Prof. Dr. Adalbert F.X. Wilhelm. \footnote{ https://www.jacobs-university.de/directory/wilhelm}

\section{Conclusion}
In our work, we show how AI could shape the future of Pneumonia Detection. We presented our model, performing the detection with 91,04\% of prediction's accuracy that could be used for early detection. However, as any other technology, AI is a multiplier for human ingenuity not a replacement for it. Therefore, in our paper, we also addressed various technical, legal, ethical, and logistical issues, with a blueprint of possible solutions, to ensure fairness and availability of the technology for everyone.


\bibliographystyle{paperstyle}

\begin{thebibliography}{}
\bibitem{1}
Our World in Data.\newblock https://ourworldindata.org/pneumonia

\bibitem{2}Kaushik, V. Nayyar, Anand. Kataria, Gaurav. Jain, Rachna.
\newblock 2020.
\newblock Pneumonia Detection Using Convolutional Neural Networks (CNNs). \newblock 471-483. 10.1007/978-981-15-3369-3\_36.

\bibitem{3}
World Health Organization.
\newblock https://www.who.int/health-topics/pneumonia

\bibitem{4}
National Heart, Lung and Blood Institute.
\newblock https://www.nhlbi.nih.gov/health-topics/pneumonia.

\bibitem{5}
Pacis, Danica Mitch M., et al.
\newblock 2018.
\newblock Trends in Telemedicine Utilizing Artificial Intelligence.
\newblock p. 040009. DOI.org (Crossref), doi:10.1063/1.5023979.

\bibitem{6}
Healthcare IT News.
\newblock https://www.healthcareitnews.com/news/86-healthcare-companies-use-some-form-ai.

\bibitem{7}
Topol, Eric J.
\newblock High-Performance Medicine: The Convergence of Human and Artificial Intelligence
\newblock Nature Medicine, vol. 25, no. 1, Jan. 2019, pp. 44–56. DOI.org (Crossref), doi:10.1038/s41591-018-0300-7.y.

\bibitem{8}
Stanford News.
\newblock Stanford algorithm can diagnose pneumonia better than radiologists.
\newblock https://news.stanford.edu/2017/11/15/algorithm-outperforms-radiologists-diagnosing-pneumonia/

\bibitem{9}
Stanford Medicine News.
\newblock Artificial intelligence rivals radiologists in screening X-rays for certain diseases.
\newblock https://med.stanford.edu/news/all-news/2018/11/ai-outperformed-radiologists-in-screening-x-rays-for-certain-diseases.html.

\bibitem{10}
Science.
\newblock Artificial intelligence could revolutionize medical care. But don’t trust it to read your x-ray just yet.
\newblock https://www.sciencemag.org/news/2019/06/artificial-intelligence-could-revolutionize-medical-care-don-t-trust-it-read-your-x-ray

\bibitem{11}
Pan American Health Organization.
\newblock World Radiography Day: Two-Thirds of the World's Population has no Access to Diagnostic Imaging.
\newblock https://www.paho.org/hq/index.php?option=com\_content\&view=article\&id=7410:2012-dia-radiografia-dos-tercios-poblacion-mundial-no-tiene-acceso-diagnostico-imagen\&Itemid=1926\&lang=en

\bibitem{12}
Kieran Murphy.
\newblock HOW DATA WILL IMPROVE HEALTHCARE WITHOUT ADDING STAFF OR BEDS.
\newblock https://www.wipo.int/edocs/pubdocs/en/wipo\_pub\_gii\_2019-chapter8.pdf

\bibitem{13}
Pranav R., Jeremy I., Kaylie Z., Brandon Y., Hershel M., Tony D., Daisy D., Aarti B., Robyn L. B., Curtis L., Katie S., Matthew P. L., Andrew Y. Ng.
\newblock 2017.
\newblock CheXNet: Radiologist-Level Pneumonia Detection on Chest X-Rays with Deep Learning.
\newblock https://arxiv.org/abs/1711.05225.

\bibitem{14}
D. Varshni, K. Thakral, L. Agarwal, R. Nijhawan and A. Mittal.
\newblock 2019.
\newblock Pneumonia Detection Using CNN based Feature Extraction.
\newblock IEEE International Conference on Electrical, Computer and Communication Technologies (ICECCT), Coimbatore, India, 2019, pp. 1-7, doi: 10.1109/ICECCT.2019.8869364.

\bibitem{15}
Khan, W. Zaki, N. Ali, L.
\newblock 2020.
\newblock Intelligent Pneumonia Identification from Chest X-Rays: A Systematic Literature Review.
\newblock Cold Spring Harbor Laboratory. https://doi.org/10.1101/2020.07.09.20150342

\bibitem{16}
Militante, S. V., and B. G. Sibbaluca.
\newblock Pneumonia Detection Using Convolutional Neural Networks.
\newblock International Journal of Scientific \& Technology Research 9, no. 04 (2020): 1332-1337.

\bibitem{17}
Internet Users in India (2020 – 21) Statistics
\newblock https://nanocms.in/internet-users-in-india/.

\bibitem{18}
X-ray (Radiography) - Chest
\newblock https://www.radiologyinfo.org/en/info.cfm?pg=chestrad.

\bibitem{19}
Han, Song, et al.
\newblock 2016.
\newblock Deep Compression: Compressing Deep Neural Networks with Pruning, Trained Quantization and Huffman Coding.
\newblock http://arxiv.org/abs/1510.00149.

\bibitem{20}
Kaggle.
\newblock Chest X-Ray Images (Pneumonia).
\newblock https://www.kaggle.com/paultimothymooney/chest-xray-pneumonia

\bibitem{21}
Ronneberger, O., Fischer, P., Brox, T.
\newblock 2015.
\newblock U-Net: Convolutional Networks for Biomedical Image Segmentation.
\newblock In Lecture Notes in Computer Science (pp. 234–241). Springer International Publishing.
\newblock https://doi.org/10.1007/978-3-319-24574-4\_28

\bibitem{22}
MIT News. Data Diversity
\newblock https://news.mit.edu/2016/variety-subsets-large-data-sets-machine-learning-1216.

\bibitem{23}
Kelly, C.J., Karthikesalingam, A., Suleyman, M. et al.
\newblock Key challenges for delivering clinical impact with artificial intelligence.
\newblock BMC Med 17, 195 (2019).
\newblock https://doi.org/10.1186/s12916-019-1426-2.

\bibitem{24}
Rodrigues, R.
\newblock 2020.
\newblock Legal and human rights issues of AI: gaps, challenges and vulnerabilities. Journal of Responsible Technology, 100005.
\newblock https://doi.org/10.1016/j.jrt.2020.100005.

\bibitem{25}
General Data Protection Regulation. GDPR.
\newblock https://gdpr-info.eu.

\bibitem{26}
CNIL.
\newblock https://www.cnil.fr/en/data-protection-around-the-world.

\bibitem{27}
UX Collective.
\newblock How usability impacts profit 
\newblock https://uxdesign.cc/the-conversion-usability-framework-3e2068edebbc.

\bibitem{28}
Special Report: Trust in Technology.
\newblock https://www.edelman.com/sites/g/files/aatuss191/files/2020-02/2020\%20Edelman20Trust\%20Barometer\%20Tech\%20Sector\%20Report\_1.pdf.

\bibitem{29}
Statista. Trust in artificial intelligence
\newblock https://www.statista.com/statistics/948531/trust-artificial-intelligence-country/.

\bibitem{30}
Our world in data.
\newblock https://ourworldindata.org/internet

\bibitem{31}
Statista. Worldwide Households With Computer.
\newblock https://www.statista.com/statistics/748551/worldwide-households-with-computer/.

\bibitem{32}
Our world in data.
\newblock https://ourworldindata.org/energy-access.

\end{thebibliography}

\end{document}